\documentclass[twoside]{article}
\usepackage[style=authoryear]{biblatex}
\usepackage{hyperref}   
\usepackage{amsfonts}       
\usepackage{nicefrac}       
\usepackage[utf8]{inputenc} 
\usepackage{graphicx}
\usepackage{bbm}
\usepackage{multirow}
\usepackage{amsmath}
\usepackage{wrapfig}
\usepackage{booktabs}

\newcommand{\MU}{\boldsymbol{\mu}}
\newcommand{\tMU}{\tilde{\boldsymbol{\mu}}}

\newcommand{\rmM}{\mathbf{M}}

\usepackage[accepted]{aistats2021}
\begin{document}

%

%

\twocolumn[

\aistatstitle{Neural Enhanced Belief Propagation on Factor Graphs}

\aistatsauthor{ Victor Garcia Satorras \And Max Welling }

\aistatsaddress{ UvA-Bosch Delta Lab \\University of Amsterdam  \\ v.garciasatorras@uva.nl \And UvA-Bosch Delta Lab \\University of Amsterdam \\ m.welling@uva.nl  } ]

\begin{abstract}
A graphical model is a structured representation of locally dependent random variables. A traditional method to reason over these random variables is to perform inference using belief propagation. When provided with the true data generating process, belief propagation can infer the optimal posterior probability estimates in tree structured factor graphs. However, in many cases we may only have access to a poor approximation of the data generating process, or we may face loops in the factor graph, leading to suboptimal estimates. In this work we first extend graph neural networks to factor graphs (FG-GNN). We then propose a new hybrid model that runs conjointly a FG-GNN with belief propagation. The FG-GNN receives as input messages from belief propagation at every inference iteration and outputs a corrected version of them. As a result, we obtain a more accurate algorithm that combines the benefits of both belief propagation and graph neural networks. We apply our ideas to error correction decoding tasks, and we show that our algorithm can outperform belief propagation for LDPC codes on bursty channels.

\end{abstract}
\section{Introduction}

Graphical models \parencite{bishop2006pattern, murphy2012machine} are a structured representation of locally dependent random variables, that combine concepts from probability and graph theory. A standard way to reason over these random variables is to perform inference on the graphical model using message passing algorithms such as Belief Propagation (BP) \parencite{pearl2014probabilistic, murphy2013loopy}. 
Provided that the true generative process of the data is given by a non-loopy graphical model, BP is guaranteed to compute the optimal (posterior) marginal probability distributions. However, in real world scenarios, we may only have access to a poor approximation of the true distribution of the graphical model, leading to sub-optimal estimates. In addition, an important limitation of belief propagation is that on graphs with loops BP computes an approximation to the desired posterior marginals or may fail to converge at all.

In this paper we present a hybrid inference model to tackle these limitations. We cast our model as a message passing method on a factor graph that combines messages from BP and from a Graph Neural Network (GNN). The GNN messages are learned from data and complement the BP messages. The GNN receives as input the messages from BP at every inference iteration and delivers as output a refined version of them back to BP. As a result, given a labeled dataset, we obtain a more accurate algorithm that outperforms either Belief Propagation or Graph Neural Networks when run in isolation in cases where Belief Propagation is not guaranteed to obtain the optimal marginals.

Belief Propagation has demonstrated empirical success in a variety of applications: Error correction decoding algorithms \parencite{mceliece1998turbo}, combinatorial optimization in particular graph coloring and satisfiability \parencite{braunstein2004survey}, inference in markov logic networks \parencite{richardson2006markov}, the Kalman Filter is a special case of the BP algorithm \parencite{yedidia2003understanding, welch1995introduction} etc. One of its most successful applications is Low Density Parity Check codes (LDPC) \parencite{gallager1962low} an error correction decoding algorithm that runs BP on a loopy bipartite graph. LDPC is currently part of the  Wi-Fi 802.11 standard, it is an optional part of 802.11n and 802.11ac, and it has been adopted for 5G, the fifth generation wireless technology that began wide deployment in 2019. Despite being a loopy algorithm, its bipartite graph is typically very sparse which reduces the number of loops and increases the cycle size. As a result, in practice LDPC has shown excellent results in error correction decoding tasks and performs close to the Shannon limit in Gaussian channels.

However, a Gaussian channel is an approximation of the more complex noise distributions we encounter in the real world. Many of these distributions have no analytical form, but we can approximate them from data. In this work we show the robustness of our algorithm over LDPC codes when we assume such a non-analytical form. Our hybrid method is able to closely match the performance of LDPC in Gaussian channels while outperforming it for deviations from this assumption (i.e. a bursty noise channel \parencite{gilbert1960capacity, kim2018communication}). 

The three main contributions in our work are: i) We extend the standard graph neural network equations to factor graphs (FG-GNN). ii) We present a new hybrid inference algorithm, Neural Enhanced Belief Propagation (NEBP) that refines BP messages using the FG-GNN. iii) We apply our method to an error correction decoding problem for a non-Gaussian (bursty) noise channel and show clear improvement on the Bit Error Rate over existing LDPC codes.

\section{Background}
\subsection{Factor Graphs}
Factor graphs \parencite{loeliger2004introduction} are a convenient way of representing graphical models. A factor graphs is a bipartite graph that interconnects a set of factors $f_s(\mathbf{x}_s)$ with a set of variables $\mathbf{x}_s$, each factor defining dependencies among its subset of variables. A global probability distribution $p(\mathbf{x})$ can be defined as the product of all these factors $p(\mathbf{x}) = \frac{1}{Z} \prod_{s \in \mathcal{F}}f_s(\mathbf{x}_s)$, where $Z$ is the normalization constant of the probability distribution. A visual representation of a Factor Graph is depicted in the left image of Figure \ref{fig:factor_graph_gnn}.

\subsection{Belief Propagation} \label{sec:belief_propagation}


Belief Propagation \parencite{bishop2006pattern}, also known as the sum-product algorithm, is a message passing algorithm that performs inference on graphical models by locally marginalizing over random variables. It exploits the structure of factor graphs, allowing more efficient computation of the marginals. Belief Propagation directly operates on factor graphs by sending messages (real valued functions) on its edges. These messages exchange \textit{beliefs} of the sender nodes about the receiver nodes, thereby transporting information about the variable's probabilities. We can distinguish two types of messages: those that go from variables to factors and those that go from factors to variables.

\textbf{Variable to factor:} $\mu_{x_m \rightarrow f_s}(x_m)$ is the product of all incoming messages to variable $x_m$ from all neighbor factors $\mathcal{N}(x_m)$ except for factor $f_s$.
\begin{equation} \label{eq:bp_variable2factor}
\mu_{x_m \rightarrow f_s}(x_m) = \prod_{l \in \mathcal{N}(x_m) \setminus f_s} \mu_{f_l \rightarrow x_m}(x_m)
\end{equation}

\textbf{Factor to variable:} $\mu_{f_s \rightarrow x_n}(x_n)$ is the product of the factor $f_s$ itself with all its incoming messages from all variable neighbor nodes except for $x_n$ marginalized over all associated variables $\mathbf{x}_s$ except $x_n$.
\begin{equation}\label{eq:bp_factor2variable} \small
\mu_{f_s \rightarrow x_n}(x_n) = \sum_{\mathbf{x}_s \backslash x_n} f_s(\mathbf{x}_s)   \prod_{m \in \mathcal{N}(f_s) \setminus n} \mu_{x_m \rightarrow f_s} (x_m)
\end{equation}

To run the Belief Propagation algorithm, messages are initialized with uniform probabilities, and the two above mentioned operations are then recursively run until convergence. One can subsequently obtain marginal estimates $p(x_n)$ by multiplying all incoming messages from the neighboring factors:

\begin{equation} \label{eq:computing_marginal}
p(x_n) \propto  \prod_{s \in \mathcal{N}(x_n)}  \mu_{f_s \rightarrow x_n}(x_n)
\end{equation}

From now on, we simplify notation by removing the argument of the messages function. In the left side of Figure \ref{fig:factor_graph_gnn} we can see the defined messages on a factor graph where black squares represent factors and blue circles represent variables.

\subsection{LDPC codes} \label{sec:ldpc}
In this paper we will apply our proposed method to error correction decoding.
 Low Density Parity Check (LDPC) codes \parencite{gallager1962low, mackay2003information} are linear codes used to correct errors in data transmitted through noisy communication channels. The sender encodes the data with redundant bits while the receiver has to decode the original message. In an LDPC code, a parity check sparse matrix $\mathbf{H} \in \mathbb{B}^{(n-k)  \times n}$ is designed, such that given a code-word $\mathbf{x} \in \mathbb{B}^n$ of $n$ bits the product of $\mathbf{H}$ and $\mathbf{x}$ is constrained to equal zero: $\mathbf{H}\mathbf{x} = 0$. $\mathbf{H}$ can be interpreted as an adjacency matrix that connects $n$ variables (i.e. the transmitted bits) with $(n-k)$ factors, i.e. the parity checks that must sum to 0. The entries of $\mathbf{H}$ are $1$ if there is an edge between a factor and a variable, where rows index factors and columns index variables. For a linear code $(n,k)$, $n$ is the number of variables and $(n-k)$ the number of factors. The prior probability of the transmitted code-word $P(\mathbf{x}) \propto \mathbbm{1}[\mathbf{H} \mathbf{x}=\mathbf{0} \bmod 2]$  can be factorized as:
\begin{equation} \label{eq:factors_ldpc_px}
P(\mathbf{x}) \propto \prod_{s} \mathbbm{1}\big[\sum_{n \in \mathcal{N}(s)} x_{n}=0 \bmod  2\Big] = \prod_{s} f_s(\mathbf{x}_s)
\end{equation}
At the receiver we get a noisy version of the code-word, $\mathbf{r}$. The noise is assumed to be i.i.d, therefore we can express the probability distribution of the received code-word as $\mathbf{x}$ as $P(\mathbf{r} | \mathbf{x}) = \prod_{n} P\big(r_{n} | x_{n} \big)$. Finally we can express the posterior distribution of the transmitted code-word given the received signal as:
\begin{equation} \label{eq:prob_x_r}
    P(\mathbf{x} | \mathbf{r}) \propto P(\mathbf{x}) P(\mathbf{r} | \mathbf{x})
\end{equation}
Equation \ref{eq:prob_x_r} is a product of factors, where some factors in $P(\mathbf{x})$ (eq. \ref{eq:factors_ldpc_px}) are connected to multiple variables expressing a constraint among them. Other factors $P(\mathbf{r} | \mathbf{x})$ are connected to a single variable expressing a prior distribution for that variable. A visual representation of this factor graph is shown in the left image of Figure \ref{fig:factor_graph_gnn}. Finally, in order to infer the transmitted code-word $\mathbf{x}$ given $\mathbf{r}$, we can just run (loopy) Belief Propagation described in section \ref{sec:belief_propagation} on the Factor Graph described above (equation \ref{eq:prob_x_r}). In other words, error correction with LDPC codes can be interpreted as an instance of Belief Propagation applied to its associated factor graph.

\section{Related Work}

One of the closest works to our method is \parencite{satorras2019combining} which also combines graphical inference with graph neural networks. However, in that work, the model is only applied to the Kalman Filter, a hidden Gaussian Markov model for time sequences, and all factor graphs are assumed to be pair-wise. In our case, we run the GNN in arbitrary Factor Graphs, and we hybridize Belief Propagation, which allows us to enhance one of its main applications (LDPC codes). Other works also learn an inference model from data like Recurrent Inference Machines \parencite{putzky2017recurrent} and Iterative Amortized Inference \parencite{marino2018iterative}. However, in our case we are presenting a hybrid algorithm instead of a fully learned one. Additionally in \parencite{putzky2017recurrent} graphical models play no role.

Our work is also related to meta learning \parencite{schmidhuber1987evolutionary, andrychowicz2016learning} in the sense that it learns a more flexible algorithm on top of an already existing one. It also has some interesting connections to the ideas from the consciousness prior \parencite{bengio2017consciousness} since our model is an actual implementation of  a sparse factor graph that encodes prior knowledge about the task to solve.

Another interesting line of research concerns the convergence of graphical models with neural networks. In \parencite{mirowski2009dynamic}, the conditional probability distributions of a graphical model are replaced with trainable factors. \parencite{johnson2016composing} learns a graphical latent representation and runs Belief Propagation on it. Combining the strengths of convolutional neural networks and conditional random fields has shown to be effective in image segmentation tasks \parencite{chen2014semantic, zheng2015conditional}. A model to run Neural Networks on factor graphs was also introduced in \parencite{zhang2019factor}. However, in our case, we simply adjust the Graph Neural Network equations to the factor graph scenario as a building block for our hybrid model (NEBP).

More closely to our work, \parencite{yoon2018inference} trains a graph neural network to estimate the marginals in Binary Markov Random Fields (Ising model) and the performance is compared with Belief Propagation for loopy graphs. In our work we are proposing a hybrid method that combines the benefits of both GNNs and BP in a single model. In \parencite{nachmani2016learning} some weights are learned in the edges of the Tanner graph for High Density Parity Check codes, in our case we use a GNN on the defined graphical model and we test our model on Low Density Parity Check codes, one of the standards in communications for error decoding. A subsequent work \parencite{liu2019neural} uses the model from \parencite{nachmani2016learning} for quantum error correcting codes. Recently,  \parencite{kuck2020belief} presented a strict generalization of Belief Propagation with Neural Networks, in contrast, our model augments Belief Propagation with a Graph Neural Network which learned messages are not constrained to the message passing scheme of Belief Propagation, refining BP messages without need to backpropagate through them.

\section{Method} \label{sec:method}

\subsection{Graph Neural Network for Factor Graphs}

\begin{figure*}[t!] 
\center

  \includegraphics[width=1.0\textwidth]{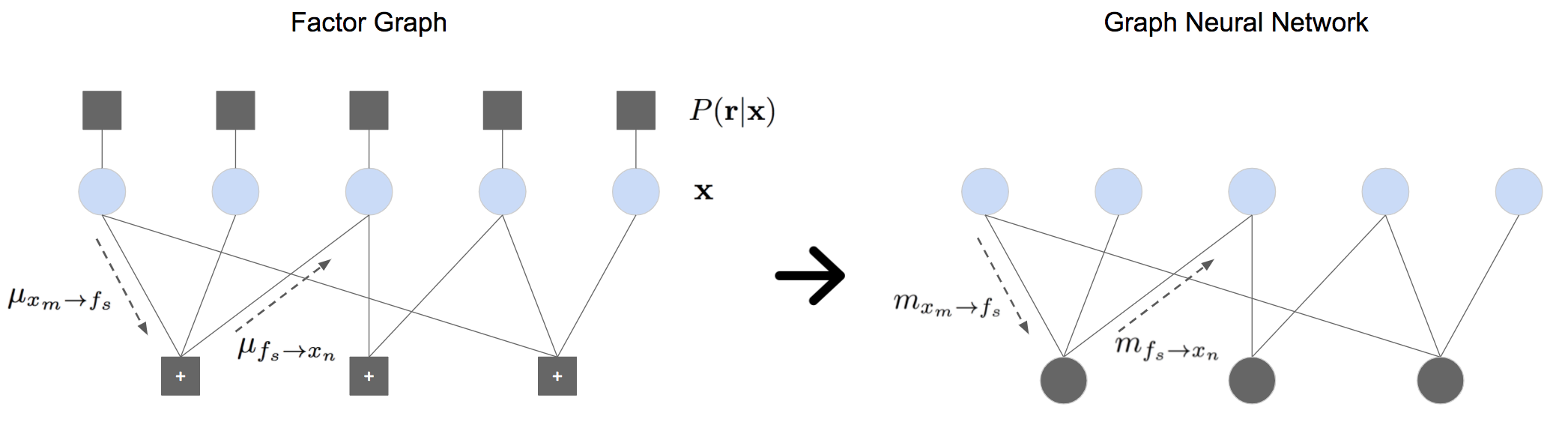}
  
  \caption{Visual representation of a LDPC Factor Graph (left) and its equivalent representation in our Graph Neural Network (right). In the Factor Graph, factors are displayed as black squares, variables as blue circles. In the Graph Neural Network, nodes associated to factors are displayed as black circles. Nodes associated to variables are displayed as blue circles.}
 \label{fig:factor_graph_gnn}
\end{figure*}
 
We will propose a hybrid method to improve Belief Propagation (BP) by combining it with Graph Neural Networks (GNNs). Both methods can be seen as message passing on a graph. However, where BP sends messages that follow directly from the definition of the graphical model, messages sent by GNNs must be learned from data. To achieve seamless integration of the two message passing algorithms, we will first extend GNNs to factor graphs.

 \begin{table} [h!] 
\begin{center} 
\begin{tabular}{ |c|c| } 
\hline
 & GNN  \\
\hline
\multirow{2}{3.5em}{$v \rightarrow e$} & \multirow{2}{13em}{
$m_{i\rightarrow j}^t = \phi_{e}(h_{i}^t, h_{j}^t, a_{ij})$}\\ 
 &  \\
\hline
\multirow{4}{3.5em}{$e \rightarrow v$} & \multirow{2}{13em}{ $m_{j}^t = \sum_{i \in \mathcal{N}(j)} m_{i\rightarrow j}^t$}   \\ 
 &   \\ \cline{2-2}

 & \multirow{2}{13em}{$h_{j}^{t+1} = \phi_{v}([m_j^t, a_{j}], h_{j}^t)$}   \\ 
 &   \\
\hline
\end{tabular} 
\end{center} \caption{Graph Neural Network equations.} \label{table:gnn}
\end{table}

 Graph Neural Networks \parencite{bruna2013spectral, defferrard2016convolutional, kipf2016semi} operate on graph-structured data by modelling interactions between pairs of nodes. A graph is defined as a tuple $\mathcal{G} = (\mathcal{V}, \mathcal{E})$, with nodes $v \in \mathcal{V}$ and edges $e \in \mathcal{E}$. Table  \ref{table:gnn} presents the edge and node operations that a GNN defines on a graph using similar notation as \parencite{gilmer2017neural}.

The message passing procedure of a GNN is divided into two main steps: from node embeddings to edge embeddings $v \rightarrow e$, and from edge to nodes $e \rightarrow v$. Where $h_i^t$ is the embedding of a node $v_i$, $\phi_e$ is the edge operation, and $m_{i\rightarrow j}^t$ is the embedding of the $e_{ij}$ edge. 
First, the edge embeddings $m_{i\rightarrow j}$ are computed, which one can interpret as \textit{messages}, next we sum all node $v_j$ incoming messages. After that, the embedding representation for node $v_j$, $h_j^{t}$, is updated through the node function $\phi_v$. Values $a_{ij}$ and $a_j$ are optional edge and node attributes respectively.

In order to integrate the GNN messages with those of BP we have to run them on a Factor Graph. In \parencite{yoon2018inference} a GNN was defined on pair-wise factor graphs (ie. a factor graph where each factor contains only two variables). In their work each variable of the factor graph represents a node in the GNN, and each factor connecting two variables represented an edge in the GNN. Properties of the factors were associated with edge attributes $a_{ij}$. The mapping between GNNs and Factor Graphs becomes less trivial when each factor may contain more than two variables. We can then no longer consider each factor as an edge of the GNN. In this work we propose special case of GNNs to run on factor graphs with an arbitrary number of variables per factor.

Similarly to Belief Propagation, we first consider a Factor Graph as a bipartite graph $\mathcal{G}_f = (\mathcal{V}_f, \mathcal{E}_f)$ with two type of nodes $\mathcal{V}_f = \mathcal{X}\cup \mathcal{F}$, variable-nodes $v_x \in \mathcal{X}$ and factor-nodes $v_f \in \mathcal{F}$, and two types of edge interactions, depending on if they go from factor-node to variable-node or vice-versa. With this graph definition, all interactions are again pair-wise (between factor-nodes and variable-nodes in the bipartite graph).

A mapping between a factor graph and the graph we use in our GNN is shown in Figure \ref{fig:factor_graph_gnn}. All factors from the factor graph are assumed to be factor-nodes in the GNN. We make an exception for factors connected to only one variable which we simply consider as attributes of that variable-node in order to avoid redundant nodes. Once we have defined our graph, we use the GNN notation from Table \ref{table:gnn}, and we re-write it specifically for this new graph. From now on we reference these new equations as FG-GNN. Equations for the FG-GNN are presented in Table \ref{table:fgnn}.

\begin{table} [h]
\begin{center} 
\begin{tabular}{ |c|c| } 
\hline
   & FG-GNN \\
\hline
\multirow{2}{3.5em}{$v \rightarrow e$}    & $m_{x\rightarrow f}^t = \phi_{x \rightarrow f}(h_f^t, h_x^t, a_{x \rightarrow f})$ \\ 
   & $m_{f\rightarrow x}^t = \phi_{f \rightarrow x}(h_x^t, h_f^t, a_{f \rightarrow x})$ \\
\hline
\multirow{4}{3.5em}{$e \rightarrow v$}    & $m_{f}^t = \sum_{x \in \mathcal{N}(f)} m_{x\rightarrow f}^t$ \\ 
   & $m_{x}^t = \sum_{f \in \mathcal{N}(x)} m_{f\rightarrow x}^t$ \\ \cline{2-2}
   & $h_f^{t+1} = \phi_{v_{f}}([m_{f}^t, a_f], h_f^t)$ \\ 
   & $h_x^{t+1} = \phi_{v_{x}}([m_{x}^t, a_v], h_x^t)$ \\
\hline
\end{tabular} 
\end{center} \caption{Graph Neural Network for a Factor Graph} \label{table:fgnn}
\end{table}

Notice that in the GNN we did not have two different kind of variables in the graph and hence we only needed one edge function $\phi_e$ (but notice that the order of the arguments of this function matters so that a message from $i\rightarrow j$ is potentially different from the message in the reverse direction). For the FG-GNN however, we now have two types of nodes, which necessitate two types of edge functions, $\phi_{x \rightarrow f}$ and $\phi_{f \rightarrow x}$, depending on whether the message was sent by a variable or a factor node. In addition, we also have two type of node embeddings $h_x$ and $h_f$ for the two types of nodes $v_x$ and $v_f$. Again we sum over all incoming messages for each node, but now in the node update we have two different functions, $\phi_{v_f}$ for the factor-nodes and $\phi_{v_x}$ for the variable-nodes. The optional edge attributes are now labeled as $a_{x \rightarrow f}$, $a_{f \rightarrow x}$, and the node attributes $a_f$ and $a_v$.

\begin{figure*}[t] 
\center
  \includegraphics[width=1.\textwidth]{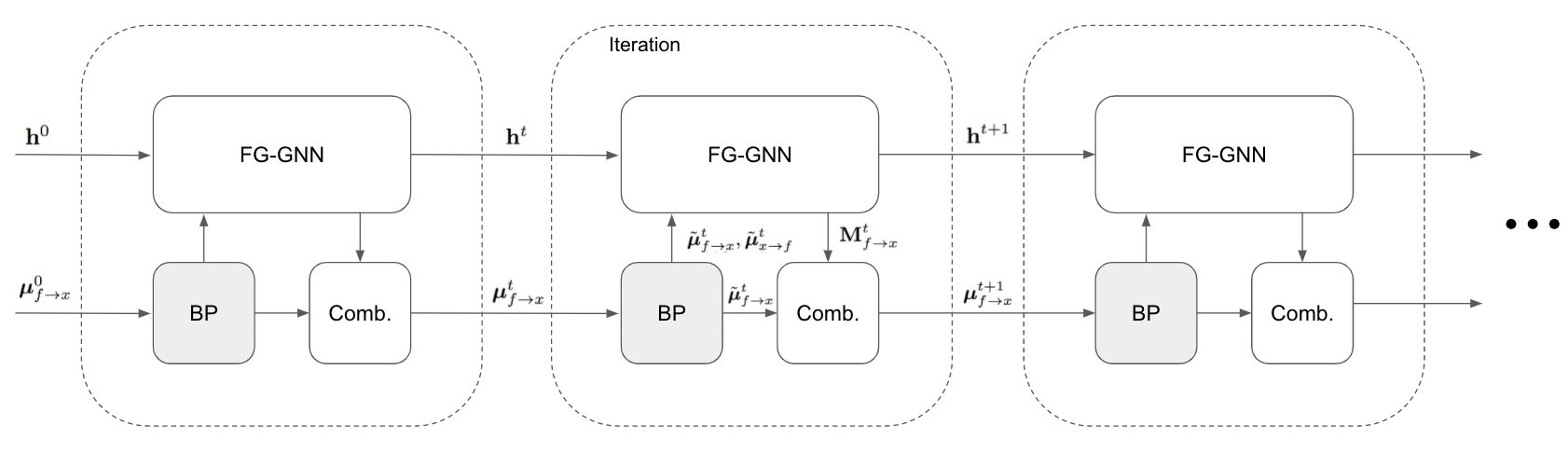}
  \caption{Graphical illustration of our Neural Enhanced Belief Propagation algorithm. Three modules are depicted in each iteration \{BP, FG-GNN, Comb.\}. Each module is associated to each one of the three lines from Equation \ref{eq:neural_enhanced_bp}.}
  \label{fig:nebp}
\end{figure*}

\subsection{Neural Enhanced Belief Propagation} \label{sec:nebp}

Now that we have defined the FG-GNN we can introduce our hybrid method that runs co-jointly with Belief Propagation on a factor graph, we denote this new method Neural Enhanced Belief Propagation (NEBP). At a high level, the procedure is as follows: after every Belief Propagation iteration, we input the BP messages into the FG-GNN. The FG-GNN then runs for two iterations and updates the BP messages. This step is repeated recursively for N iterations. After that, we can compute the marginals from the refined BP messages.

We first define the two functions $\text{BP}(\cdot)$ and $\text{FG-GNN}(\cdot)$. $\text{BP}(\cdot)$ takes as input the factor-to-node messages $\MU_{f \rightarrow x}^t$, then runs the two BP updates eqns. \ref{eq:bp_variable2factor} and \ref{eq:bp_factor2variable} respectively and outputs the result of that computation as $\tMU_{f \rightarrow x}^t, \tMU_{x \rightarrow f}^t$. We initialize $\MU_{f \rightarrow x}^{0}$ as uniform distributions.

The function $\text{FG-GNN}(\cdot)$ runs the equations displayed in Table \ref{table:fgnn}. At every $t$ iteration we give it as input the quantities $\mathbf{h}^t=\{h_x^t | x\in \mathcal{X}  \} \cup \{h_f^t | f\in \mathcal{F}  \}$, $\mathbf{a}_{x\rightarrow f}$, $\mathbf{a}_{f \rightarrow x}$ and $\mathbf{a}_v$. $\mathbf{h}^t$ is initialized randomly as $\mathbf{h}^{0}$ by sampling from a normal distribution. Moreover, the attributes $\mathbf{a}_{x\rightarrow f}$ and $\mathbf{a}_{f \rightarrow x}$ are provided to the function $\text{FG-GNN}(\cdot)$ as the messages $\tMU_{x \rightarrow f}
^t$ and $\tMU_{f \rightarrow x}
^t$ obtained from $\text{BP}(\cdot)$, as an exception, the subset of messages from $\tMU_{f \rightarrow x}$ that go from a singleton factor $f_l$ to its neighbor variable are treated as attributes $\mathbf{a}_v$. The outputs of $\text{FG-GNN}(\cdot)$ are the updated latent vectors $\mathbf{h}_x^{t+1}$ and the latent messages $\mathbf{m}_{f \rightarrow x}^t$ computed as part of the FG-GNN algorithm in Table \ref{table:fgnn}. These latent representations $\rmM_{f \rightarrow x}^{t}= \mathbf{m}_{f \rightarrow x}^{t} \cup \mathbf{h}_x^{t+1}$ will be used to update the current message estimates $\tMU_{f \rightarrow x}^t$, specifically $\mathbf{h}_x^{t+1}$ will refine those messages $\tMU_{f_l \rightarrow x}^t$ that go from a singleton factor to its neighbor variable and $\mathbf{m}_{f \rightarrow x}^{t}$ will refine the rest of messages $\tMU_{f \rightarrow x}^t \backslash \tMU_{f_l \rightarrow x}^t$. All other variables computed inside $\text{FG-GNN}(\cdot)$ are kept internal to this function.

Finally, $f_s(\cdot)$ and $f_u(\cdot)$ take as input the embeddings $\rmM_{f \rightarrow x}^{t}$ and output a refinement for the current message estimates $\tilde{\MU}_{f \rightarrow x}^{t}$. Particularly, $f_s(\cdot)$ outputs a positive scalar value that multiplies the current estimate, and $f_u(\cdot)$ outputs a positive vector which is summed to the estimate. Both functions encompass two Multi Layer Perceptrons (MLP), one MLP takes as input the node embeddings $\mathbf{h}_x^{t+1}$, and outputs the refinement for the singleton factor messages, the second MLP takes as input the edge embeddings $\mathbf{m}_{f \rightarrow x}^{t}$ and outputs a refinement for the rest of messages $\tMU_{f \rightarrow x}^t \backslash \tMU_{f_l \rightarrow x}^t$. In summary, the hybrid algorithm thus looks as follows:
    \begin{equation}\label{eq:neural_enhanced_bp} \footnotesize
\begin{array}{rlrl}
\tMU_{f \rightarrow x}^{t}, \tMU_{x \rightarrow f}^{t} &
= \mathrm{BP}\big( \MU_{f \rightarrow x}^{t} \big)\\

\rmM_{f \rightarrow x}^{t} &
= \text{FG-GNN}\big(\mathbf{h}^{t}, \tMU_{f \rightarrow x}^{t}, \tMU_{x \rightarrow f}^{t}\big)  &\\

\MU_{f \rightarrow x}^{t + 1} &
=  \tMU_{f \rightarrow x}^{t}f_{s}(\mathbf{M}_{f \rightarrow x}^{t}) + f_{u}(\mathbf{M}_{f \rightarrow x}^{t}) &

\end{array}
\end{equation}
After running the algorithm for N iterations. We obtain the estimate $\hat{p}(x_i)$ by using the same operation as in Belief Propagation (eq. \ref{eq:computing_marginal}), which amounts to taking a product of all incoming messages to node $x_i$, i.e. $\hat{p}(x_i) \propto \prod_{s\in \mathcal{N}(x_i)} \mu_{f_s \rightarrow x_i}$. From these marginal distributions we can compute any desired quantity on a node.

\subsection{Training and Loss}
The loss is computed from the estimated marginals $\hat{p}(x)$ and ground truth values $x_{gt}$, which we assume known during training. In the LDPC experiment the ground truths $x_{gt}$ are the transmitted bits which are known by the receiver during the training stage.
\begin{equation}
    \text{Loss}(\Theta)=\mathcal{L}\left(x_{gt}, \hat{p}(x)\right) + \mathcal{R}
\end{equation}
During training we back-propagate through the whole multi-layer estimation model (with each layer an iteration of the hybrid model), updating the FG-GNN, $f_s{}(\cdot)$ and $f_u{}(\cdot)$ weights $\Theta$. The number of training iterations is chosen by cross-validating. In our experiments we use the binary cross entropy loss for $\mathcal{L}$. The regularization term $\mathcal{R}$ is the mean of $f_u(\cdot)$ outputs, i.e. $\mathcal{R} = \frac{1}{N}\sum_{t} \mathrm{mean}\big( f_{u}(\mathbf{M}_{f \rightarrow x}^{t}) \big)$. It encourages the model to behave closer to Belief Propagation. In case, $f_u(\cdot)$ output is set to 0, the hybrid algorithm would be equivalent to Belief Propagation. This happens because $f_s(\cdot)$ outputs a scalar value that on its own only modifies the norm of $\tMU_{f \rightarrow x}^{t}$. Belief Propagation can operate on unnormalized beliefs, although it is a common practice to normalize them at every $\mathrm{BP}(\cdot)$ iteration to avoid numerical instabilities. Therefore, only modifying the norm of the messages on its own doesn't change our hybrid algorithm because messages are being normalized at every BP iteration.

\section{Experiments}

We analyze the performance of Belief Propagation, FG-GNNs, and our Neural Enhanced Belief Propagation (NEBP) in an error correction task where Belief Propagation is also known as LDPC (Section \ref{sec:ldpc}) and in an inference task on the Ising model (Section \ref{sec:ising_model}). In both FG-GNN and NEBP, the edge operations $\sigma_{x\rightarrow f}$, $\sigma_{f\rightarrow x}$ defined in Section \ref{sec:method} consist of two layers Multilayer Perceptrons (MLP). The node update functions $\sigma_{v_f}$ and $\sigma_{v_x}$ consist of two layer MLPs followed by a Gated Recurrent Unit \parencite{chung2014empirical}. Functions $f_s(\cdot)$ and $f_u(\cdot)$ from the NEBP combination module also contain two layers MLPs.

\subsection{Low Density Parity Check codes}

LDPC codes, explained in section \ref{sec:ldpc} are a particular case of Belief Propagation run on a bipartite graph for error correction decoding tasks. Bipartite graphs contain cycles, hence Belief Propagation is no longer guaranteed to converge nor to provide the optimal estimates. Despite this lack of guarantees, LDPC has shown excellent results near the Shannon limit \parencite{mackay1996near} for Gaussian channels. LDPC assumes a channel with an analytical solution, commonly a Gaussian channel. In real world scenarios, the channel may differ from Gaussian or it may not even have a clean analytical solution to run Belief Propagation on, leading to sub-optimal estimates. An advantage of neural networks is that, in such cases, they can learn a decoding algorithm from data. 

In this experiment we consider the bursty noisy channel from \parencite{kim2018communication}, where a signal $x_i$ is transmitted through a standard Gaussian channel $z_i \sim \mathcal{N}(0, \sigma_c^2)$, however this time, a larger noise signal $w_i \sim \mathcal{N}(0, \sigma_b^2)$ is added with a small probability $\rho$. More formally:
\begin{equation} \label{eq:bursty_channel}
    r_{i}=x_{i}+z_{i}+ p_iw_{i}
\end{equation}
Where $r_i$ is the received signal, and $p_i$ follows a Bernoulli distribution such that $p_i=1$ with probability $\rho$, and $p_i=0$ with probability $1 - \rho$. In our experiments, we set $\rho=0.05$ as done in \parencite{kim2018communication}. This bursty channel describes how unexpected signals may cause interference in the middle of a transmitted frame. For example, radars may cause bursty interference in wireless communications. In LDPC, the SNR is assumed to be known and fixed for a given frame, yet, in practice it needs to be estimated with a known preamble (the pilot sequence) transmitted before the frame. If bursty noise occurs in the middle of the transmission, the estimated SNR is blind to this new noise level.

\textbf{Dataset:} We use the parity check matrix $\mathbf{H}$ "96.3.963" from \parencite{mackay2009david} for all experiments, with $n=96$ variables and $k=48$ factors, i.e. a transmitted code-word $\mathbf{x} \in \mathbb{B}^{n}$ contains 96 bits. The training dataset consists of pairs of received and transmitted code-words $\{(\mathbf{r}_d, \mathbf{x}_d)\}_{1 \leq d \leq L}$. The transmitted code-words $\mathbf{x}$ are used as ground truth for training the decoding algorithm. The received code-words $\textbf{r}$ are obtained by transmitting $\mathbf{x}$ through the bursty channel from Equation \ref{eq:bursty_channel}. We generate samples for SNR$_{db}$ = \{0, 1, 2, 3, 4\}. Regarding the bursty noise $\sigma_b$, we randomly sample its standard deviation from a uniform distribution $\sigma_b \in [0, 5]$. We generate a validation partition of 750 code-words (150 code-words per SNR$_{db}$ value). For the training partition we keep generating samples until convergence, i.e. until we do not see further improvement in the validation accuracy.

\begin{figure*}[t] 
\center
  \includegraphics[width=1.0\textwidth]{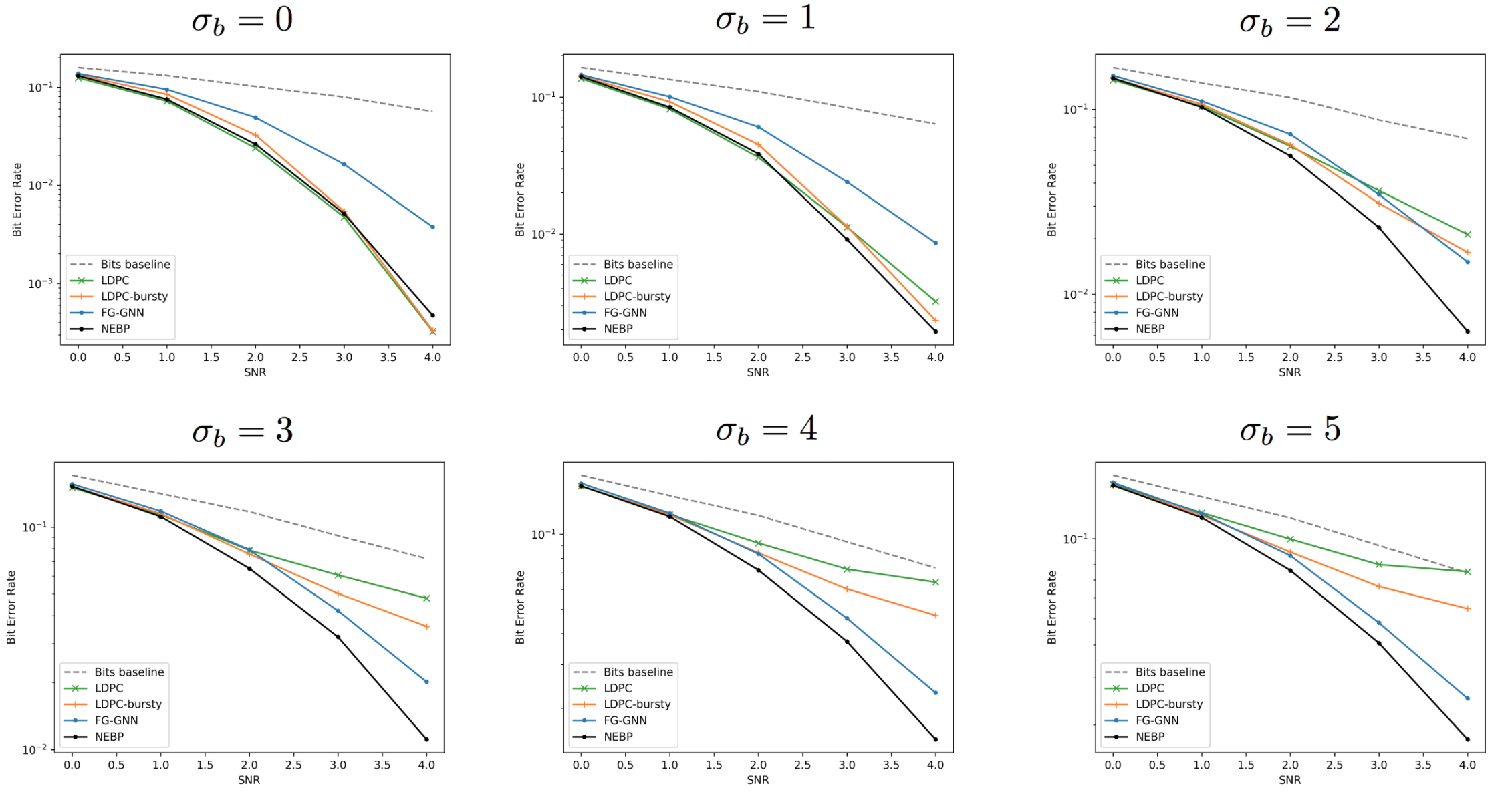}
  \caption{Bit Error Rate (BER) with respect to the Signal to Noise Ratio (SNR) for different bursty noise values $\sigma_b \in \{0, 1, 2, 3, 4, 5\}$.}
\label{fig:ldpc_experiment}
\vspace{-6pt}
\end{figure*}

\textbf{Training procedure:} We provide as input to the model the received code-word $\mathbf{r}_d$ and the SNR for that code-word. These values are provided as node attributes $a_v$ described in Section \ref{sec:method}. We run the algorithms for 20 iterations and the loss is computed as the cross entropy between the estimated $\hat{\mathbf{x}}$ and the ground truth $\mathbf{x}_d$. We use an Adam optimizer \parencite{kingma2014adam} with a learning rate $2e^{-4}$ and batch size of 1. The number of hidden features is 32 and all activation functions are 'Selus' \parencite{klambauer2017self}. As a evaluation metric we compute the Bit Error Rate (BER), which is the number of error bits divided by the total amount of transmitted bits. The number of test code-words we used to evaluate each point from our plots (Figure \ref{fig:ldpc_experiment}) is at least $\frac{200}{\hat{\text{BER}} \cdot n}$, where $n$ is the number of bits per code-word and $\hat{\text{BER}}$ is the estimated Bit Error Rate for that point.

\textbf{Baselines:} Beside the already mentioned methods (FG-GNN and standard LDPC error correction decoding), we also run two extra baselines. The first one we call \textit{Bits baseline}, which consists of independently estimating each bit that maximizes $p(r_i | x_i)$. The other baseline, called \textit{LDPC-bursty}, is a variation of LDPC, where instead of considering a SNR with a noise level $\sigma_c^2 = \text{var}[z]$, we consider the noise distribution from Equation \ref{eq:bursty_channel} such that now the noise variance is $\sigma^2 = \text{var}[z + pw] =  \sigma_c^2 + (\rho(1-\rho) + \rho^2)E_{\sigma_b^2}[\sigma_b^2]$. This is a fairer comparison to our learned methods, because even if we are blind to the $\sigma_b$ value, we know there may be a bursty noise with probability $\rho$ and $\sigma_b \sim \mathcal{U}(0, 5)$.

\begin{figure}[h!] 
    \centering
    \includegraphics[width=0.4\textwidth]{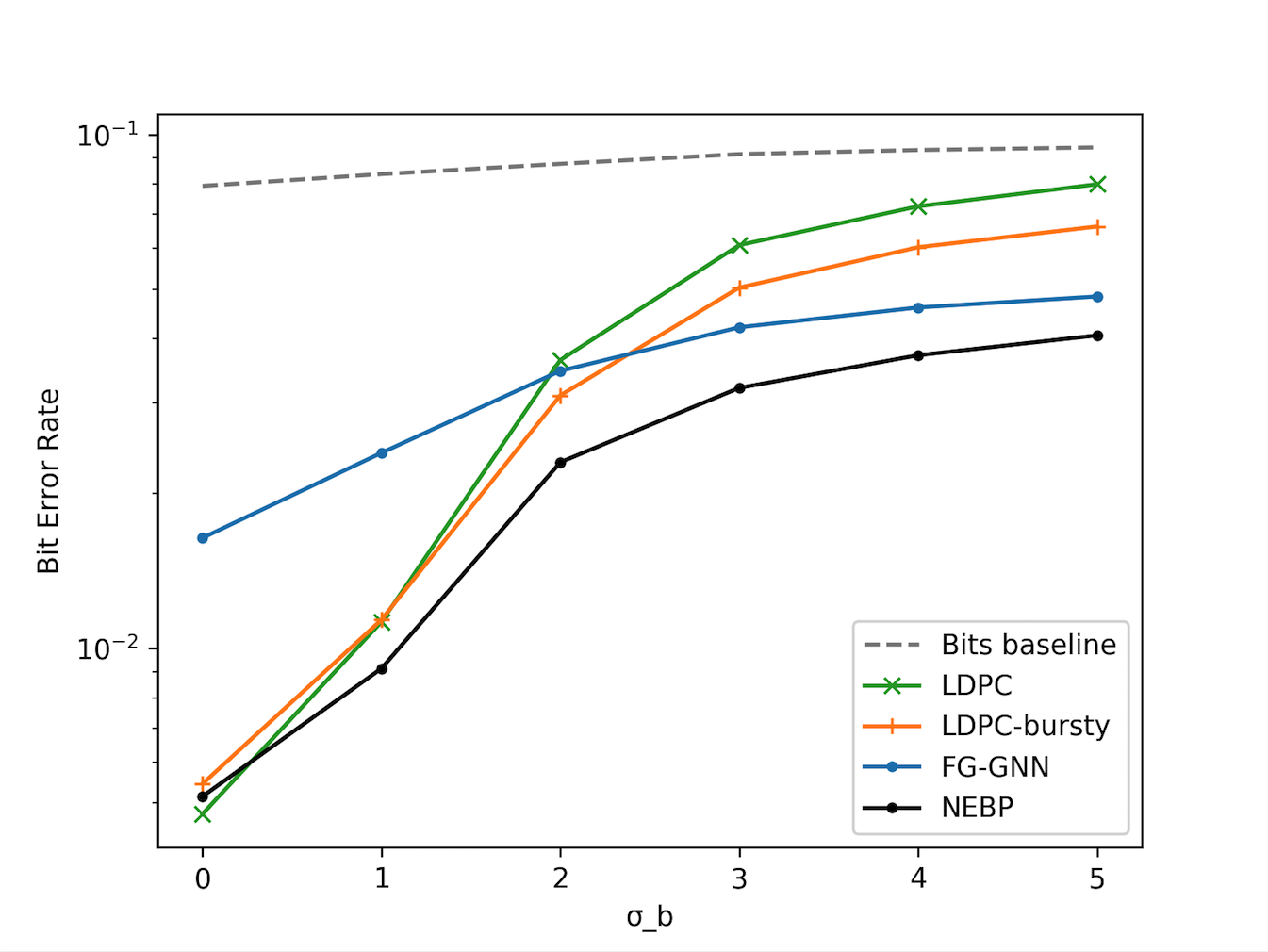}
    \caption{Bit Error Rate (BER) with respect to $\sigma_b$ value for a fixed SNR=3.}
    \vspace{-4pt}
    \label{fig:ldpc_plot2}
\end{figure}

\begin{table*}[t]
\small
\begin{center}
\begin{tabular}{ l c c c c}
\toprule
 \textbf{Model} & True Graphical Model $(u=0)$ & Mismatch $(u=0.2)$ & Mismatch $(u=0.4)$& Mismatch $(u=0.8)$\\ 
 \midrule 
  FG-GNN & 0.0141 & 0.0570 & 0.1170 &  0.1659\\ 
 BP & 0.0190 & 0.0711 & 0.2081 & 0.3121\\ 
  BP (damping) & 0.0055 & 0.0519 & 0.1318 & 0.1961\\ 
  NEBP & 0.0091 & 0.0509 & 0.1057 & 0.1697 \\ 
 \bottomrule
\end{tabular} 
\caption{\small KL divergence between the true marginals p(x) and the estimated marginals for the Ising model dataset.}
\label{table:ising_results}
\end{center}
\end{table*}

\textbf{Results:}  In Figure \ref{fig:ldpc_experiment} we show six different plots for each of the $\sigma_b$ values \{0, 1, 2, 3, 4, 5\}. In each plot we sweep the SNR from 0 to 4. Notice that for $\sigma_b=0$ the bursty noise is non-existent and the channel is equivalent to an Additive White Gaussian Noise channel (AWGN). LDPC has analytically been designed for this channel obtaining its best performance here. The aim of our algorithm is to outperform LDPC for $\sigma_b > 0 $ while still matching its performance for $\sigma_b=0$. As shown in the plots, as $\sigma_b$ increases, the performance of NEBP and FG-GNN improves compared to the other methods, with NEBP always achieving the best performance, and getting close to the LDPC performance for the AWGN channel ($\sigma_b=0$). In summary, the hybrid method is more robust than LDPC, obtaining competitive results to LDPC for AWGN channels but still outperforming it when bursty interferences are present. The FG-GNN instead, obtains relatively poor performance compared to LDPC for small $\sigma_b$, demonstrating that belief propagation is still a very powerful tool compared to pure learned inference for this task. Our NEBP is able to combine the benefits from both LDPC and the FG-GNN to achieve the best performance, exploiting the adaptability of FG-GNN and the prior knowledge of Belief Propagation. Finally, LDPC-bursty shows a more robust performance as we increase $\sigma_b$ but it is significantly outperformed by NEBP in bursty channels, and it also performs slightly worse than LDPC for the AWGN channel ($\sigma_b=0$).

In order to better visualize the decrease in performance as the burst variance increases, we sweep over different $\sigma_b$ values for a fixed SNR=3. The result is shown in Figure \ref{fig:ldpc_plot2}. The larger $\sigma_b$, the larger the BER. However, the performance decreases much less for our NEBP method than for LDPC and LDPC-bursty. In other words, NEBP is more robust as we move away from the AWGN assumption. We want to emphasize that in real world scenarios, the channel may always deviate from gaussian. Even if assuming an AWGN channel, its parameters (SNR) must be estimated in real scenarios. This potential deviations make hybrid methods a very promising approach.

\subsection{Ising model} \label{sec:ising_model}
In this section we evaluate our algorithm in a Binary Markov Random field, specifically the Ising model. We consider a squared lattice type Ising model defined by the following energy function $p(\mathbf{x})=      \frac{1}{Z} \exp (\mathbf{b} \cdot \mathbf{x}+\mathbf{x} \cdot \mathbf{J} \cdot \mathbf{x})$,  with variables $\mathbf{x} \in\{+1,-1\}^{|\mathcal{V}|}$. Where $\mathbf{b}$ biases individual variables and $\mathbf{J}$ couples pairs of neighbor variables, $b_i \sim \mathcal{N}(0, 0.25^2)$, $J_{ij} \sim \mathcal{N}(0, 1)$ and $|\mathcal{V}|$ is the number of variables which is set to 16. This energy function can be equally expressed as a product of singleton factors $f_i(x_i) = e^{x_i b_i}$ and pairwise factors $f_{ij}(x_i, x_j) = e^{J_{ij}x_ix_j}$. The task is to obtain the marginal probabilities $p(x_i)$ given the factor graph $p(\mathbf{x})$. As in the LDPC experiment, we assume some hidden dynamics not defined in the graphical model that difficult the inference task, we do that by adding a soft interaction among all pairs of variables sampled from a uniform probability distribution  $u_{ij} \sim \mathcal{U}(0, u)$. 
The Ising model is a loopy graphical model where the performance of Belief Propagation may significantly degrade due to cyclic information. Therefore, in this experiment we include a stronger baseline where Belief Propagation messages are damped to reduce the effect of cyclic information \parencite{koller2009probabilistic}. This modification significantly increased its performance. In contrast, in the previous experiment (LDPC), damping didn't lead to improvements. This is coherent with the fact that LDPC graphs are very sparse which minimizes cyclic information and Belief Propagation performs optimally for non cyclic graphs.

\textbf{Implementation details}: Belief Propagation takes the previously defined factors $f_i$ and $f_{ij}$ as the input factors $f_s$ defined in Section \ref{sec:belief_propagation}. In the FG-GNN baseline, values $b_i$ and $J_{ij}$ are inputted as variable and factor attributes $a_v$ and $a_f$ from Table \ref{table:fgnn} respectively. Finally, our NEBP combines FG-GNN and BP messages as explained in Section \ref{sec:nebp}. The damping parameter was chosen by cross-validating on the validation partition. All algorithms are run for 10 iterations. Results have been averaged over three runs.  All methods have been trained for 400 epochs, Adam optimizer, batch size 1 and learning rate $1e^{-5}$. All hidden layers have 64 neurons and the 'Leaky Relu' \parencite{xu2015empirical} as activation function.

\textbf{Results:} In Table \ref{table:ising_results} we present the KL divergence between the estimated marginals and the ground truth $p(x)$ for each model trained for different $u$ values. Since we are working with cyclic data, BP is not an exact inference algorithm even when provided with the true graphical model ($u=0$). Damping decreases the cyclic information such that for this setting ($u=0$) Damping Belief Propagation gives the best performance. NEBP gets a significant margin w.r.t. BP and FG-GNN for ($u=0$) and it outperforms other methods when the mismatch $u$ increases. As $u$ gets larger, we reach a point (i.e. $u=0.8$) where BP messages are no longer beneficial such that the learned messages (FG-GNN) alone perform best while we still get a close performance with NEBP.
\section{Conclusions}

In this work, we presented a hybrid inference method that enhances Belief Propagation by co-jointly running a Graph Neural Network that we designed for factor graphs. In cases where the data generating process is not fully known (e.g. the parameters of the graphical model need to be estimated from data), belief propagation doesn't perform optimally. Our hybrid model in contrast is able to combine the prior knowledge encoded in the graphical model (albeit with the wrong parameters) and combine this with a (factor) graph neural network with its parameters learned from labeled data on a representative distribution of channels. Note that we can think of this as meta-learning because the FG-GNN is not trained on one specific channel but on a distribution of channels and therefore must perform well on any channel sampled from this distribution without knowing its specific parameters. We tested our ideas on a state-of-the-art LDPC implementation with realistic bursty noise distributions. Our experiments clearly show that the neural enhancement of LDPC improves performance both relative to LDPC and relative to FG-GNN as the variance in the bursts gets larger. 


\printbibliography

\end{document}